\documentclass[letterpaper, 10 pt, conference]{ieeeconf}  % Comment this line out if you need a4paper

\IEEEoverridecommandlockouts                              % This command is only needed if 
\usepackage{graphics} % for pdf, bitmapped graphics files
\usepackage{epsfig} % for postscript graphics files

\usepackage{color}
\usepackage{url}
\usepackage{hyperref}
\usepackage{amsmath}
\usepackage{relsize}

\newcommand{\etal}{\textit{et al}.}

\definecolor{patrick_color}{rgb}{.6,.4,.05}
\definecolor{charlie_color}{rgb}{0,0,0.8}
\definecolor{jeremy_color}{rgb}{0,.7,.7}

\newcommand{\method}[0]{VFTS}
\newcommand{\network}[0]{VFTS-Net}

\title{\LARGE \bf Force/Torque Sensing for Soft Grippers using an External Camera}
\author{Jeremy A. Collins$^{1}$, Patrick Grady$^{1}$, Charles C. Kemp$^{1}$% <-this % stops a space
\thanks{$^{1}$Jeremy A. Collins, Patrick Grady, and Charles C. Kemp are with the Institute for Robotics and Intelligent Machines at the Georgia Institute of Technology (GT). This work was supported in part by NSF Award \# 2024444. Charles C. Kemp is an associate professor at GT. He also owns equity in and works part-time for Hello Robot Inc., which sells the Stretch RE1. He receives royalties from GT for sales of the Stretch RE1.}
}

\newenvironment{first_caption}
  {\par\footnotesize}
  {\par\addvspace{\bigskipamount}}

\begin{document}

\twocolumn[{%
\renewcommand\twocolumn[1][]{#1}%
\maketitle
\thispagestyle{empty}
\pagestyle{empty}

\begin{center}
\centering
\vspace{-2mm}
\includegraphics[width=0.99\linewidth]{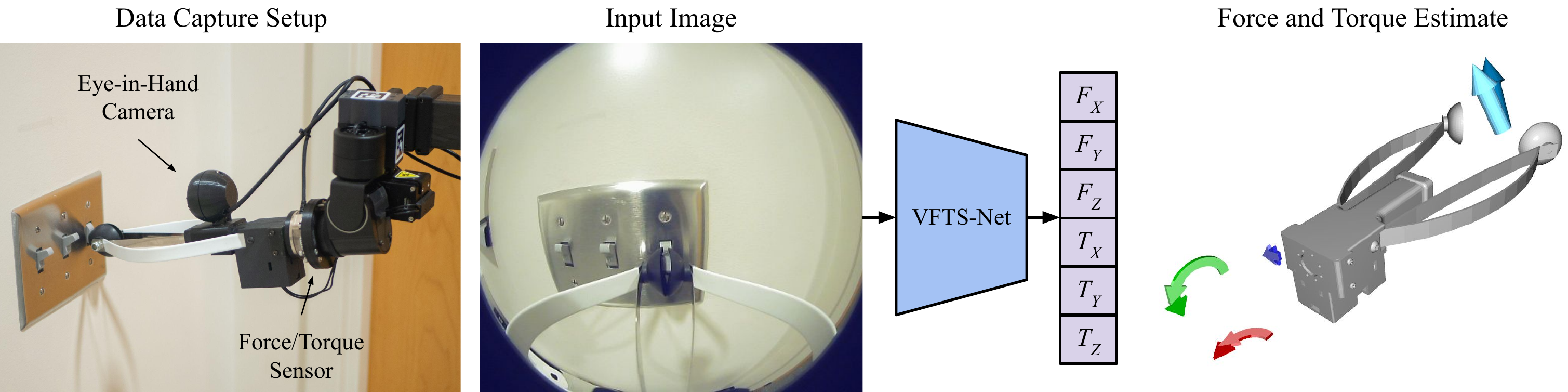}
\end{center}
\begin{first_caption}
 Fig. 1. We modify a soft robotic gripper by adding an eye-in-hand camera and a force/torque sensor. Data is collected by teleoperating the robot in a variety of home and office settings. We train a network, \network{}, to take images from the camera as input and output 3-axis forces and 3-axis torques. Estimates from \network{} are visualized as lightly shaded arrows, and ground truth measurements from the force/torque sensor are darkly shaded arrows.
\end{first_caption}
}]

\setcounter{figure}{1}  % start figure numbers at 2
\setcounter{footnote}{1}

% \maketitle
% \thispagestyle{empty}
% \pagestyle{empty}

\footnotetext{Jeremy A. Collins, Patrick Grady, and Charles C. Kemp are with the Institute for Robotics and Intelligent Machines at the Georgia Institute of Technology (GT). This work was supported by NSF Award \# 2024444. Code, data, and models are available at \url{https://github.com/Healthcare-Robotics/visual-force-torque}. Charles C. Kemp is an associate professor at GT. He also owns equity in and works part-time for Hello Robot Inc., which sells the Stretch RE1. He receives royalties from GT for sales of the Stretch RE1.}

%%%%%%%%%%%%%%%%%%%%%%%%%%%%%%%%%%%%%%%%%%%%%%%%%%%%%%%%%%%%%%%%%%%%%%%%%%%%%%%%
\begin{abstract}
Robotic manipulation can benefit from wrist-mounted force/torque (F/T) sensors, but conventional F/T sensors can be expensive, difficult to install, and damaged by high loads. We present Visual Force/Torque Sensing (\method{}), a method that visually estimates the 6-axis F/T measurement that would be reported by a conventional F/T sensor. In contrast to approaches that sense loads using internal cameras placed behind soft exterior surfaces, our approach uses an external camera with a fisheye lens that observes a soft gripper. \method{} includes a deep learning model that takes a single RGB image as input and outputs a 6-axis F/T estimate. We trained the model with sensor data collected while teleoperating a robot (Stretch RE1 from Hello Robot Inc.) to perform manipulation tasks. \method{} outperformed F/T estimates based on motor currents, generalized to a novel home environment, and supported three autonomous tasks relevant to healthcare: grasping a blanket, pulling a blanket over a manikin, and cleaning a manikin's limbs. \method{} also performed well with a manually operated pneumatic gripper. Overall, our results suggest that an external camera observing a soft gripper can perform useful visual force/torque sensing for a variety of manipulation tasks. 
\end{abstract}

%Overall, our results show that useful F/T estimates can be obtained from visual observations of soft grippers, and \method{} provides a feasible alternative to conventional wrist-mounted F/T sensors and motor-current F/T estimation. 

%Overall, our results show that useful F/T estimates can be obtained from visual observations of soft grippers operating in the wild, and \method{} provides a feasible alternative to conventional wrist-mounted F/T sensors and motor-current F/T estimation. 
% \jeremy{Vision-based Deep Estimation of Force and Torque (V-DEFT)?}
\section{INTRODUCTION}

\begin{figure*}
  \centering
  % \vspace{2mm}
  \includegraphics[width=0.99\linewidth]{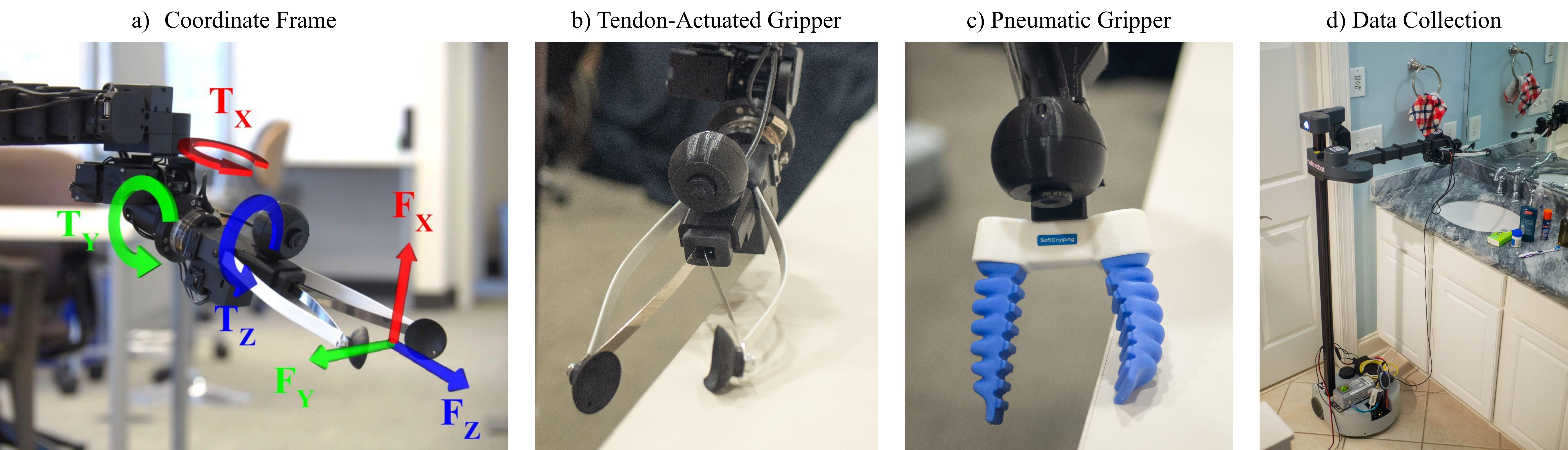}
  % \vspace{-5mm}
  \caption{a) The right-handed coordinate frame used in this paper is shown. Torques are drawn as curved arrows, while forces are drawn as straight arrows. We use the colors R/G/B to denote the X/Y/Z axes, respectively. b) Under the application of force, the tendon-actuated gripper flexures and fingertip deform against the surface. c) The fingers of the pneumatic gripper are highly compliant and deform uniformly under the application of force. d) We collected data for the tendon-actuated gripper by teleoperating the robot in a variety of settings, including a real home.}
  \label{fig:axes}
\end{figure*}

During robotic manipulation, grippers often apply forces and torques to the environment. Sensing the force and torque applied by the gripper has been useful for autonomous manipulation, but sensors that provide this information have limitations. Notably, conventional F/T sensors can be expensive, difficult to mount, and damaged by high loads. 

%and sensitive to environmental conditions such as the ambient temperature. 

For example, a common approach to measuring the load applied to a gripper is to mount an F/T sensor between the gripper and the robot's wrist. F/T sensors often use strain gauges to sense tiny deformations in an elastic element of the sensor.
%F/T sensors often rely on strain gauges placed between two rigid structures used to mount the sensor.
This approach requires that the strain gauges be resilient to the external load applied to the gripper as well as gravitational and inertial forces from the gripper itself. For many applications, the strain gauges need to be both stiff and sensitive, and protective coverings could reduce performance by interfering with the load on the strain gauges. Together, these design objectives are difficult to achieve.

We present an alternative to conventional F/T sensors. Instead of relying on the deformation of internal components, \method{} directly observes the deformation of a soft gripper using an external camera.
%We present an alternative to conventional F/T sensors which rely on the deformation of internal components. Instead, \method{} observes the deformation of the soft gripper using an external camera. 
The high compliance of soft grippers results in deformations that can be visually observed using a commodity camera. By observing this load-dependent phenomenon, the causative forces and torques can be estimated. We rigidly mount a camera with a fisheye lens to the gripper (i.e., an eye-in-hand camera). We then train a convolutional neural network, \network{}, to estimate the applied force and torque based on a single RGB image from this camera (Figure 1).

%\ref{fig:headliner}).

% \ck{298 results for "commodity camera" on Google Scholar. 4 results for "store bought camera". 1,550 for "off-the-shelf camera".}
% \jeremy{456k results for "commercially available camera"}
% \ck{the problem is that there are some very expensive commercially available cameras used for machine vision, so we need to somehow convey that it is a low-cost camera. how much did it cost?}
% \jeremy{\$57}

In contrast with conventional F/T sensors, our approach relies on a low-cost USB camera ($\sim\$60$). Our method eases installation by allowing the camera to be mounted to the exterior of the gripper rather than between the gripper and the wrist. Since the camera visually senses the loads from a distance, it is also less likely to be damaged by high loads.

Researchers have investigated related methods that involve placing a camera inside a gripper behind a compliant surface. Loads applied can be estimated by observing deformation in the surface. %, which is sometimes patterned to facilitate estimation.
In contrast, our approach uses an external camera to observe a soft gripper. Our approach does not require modification of the gripper's contact surfaces or interior. The global view of the external camera facilitates estimation of the total force and torque applied to the soft gripper. %\patrick{omit: For example, the external camera simultaneously views multiple compliant elements, such as the two spring-steel fingers with rubber fingertips used by the tendon-actuated gripper in our study. }

We provide evidence for the feasibility of our approach by collecting a dataset of in-the-wild robotic manipulation in multiple environments and testing on held-out data from novel environments with manipulation of unseen objects. We also provide an analysis indicating that lower performance corresponds with types of gripper deformation that are more difficult to visually observe. 

\method{} outperformed a baseline method that uses motor currents to estimate 6-axis F/T measurements. We also provide evidence that the estimates from \method{} can support autonomous manipulation by enabling a mobile manipulator to perform three autonomous tasks. We additionally show that \network{} can be trained to work with two distinct soft grippers: a tendon-actuated gripper and a pneumatic gripper.

%To demonstrate the applicability of our method, we collect data with two different models of soft gripper. Training data for \network{} is collected while the grippers are teleoperated or manually operated in a variety of environments. A physical F/T sensor is used to generate ground-truth measurements of force and torque. However, at test time, this sensor is not required. We collect data in a variety of indoor settings and test the performance of the network in held-out environments. Additionally, we test the real-world ability of \method{} to accomplish tasks leveraging the estimated forces and torques. We show that our method can be used for a fabric manipulation task and a human bed-bathing task. 

In summary, our paper includes the following contributions:
% \ck{revise these?}
\begin{itemize}
    % \item We present Visual Force/Torque Sensing (\method{}), a method to estimate the resultant forces and torques exerted on a soft robotic gripper using an eye-in-hand camera and a convolutional neural network.
    \item We present Visual Force/Torque Sensing (\method{}), a method that uses a convolutional neural network to estimate the forces and torques exerted on a soft gripper given an RGB image from an eye-in-hand camera.
    \item We demonstrate the utility of \method{} for real-world applications with a series of robotic tasks conducted with a mobile manipulator.
    \item We will release our code, data, and trained models.
\end{itemize}
\section{RELATED WORKS}

\subsection{Tactile Sensing using Physical Sensors}

A wide range of methods have been proposed for robotic tactile sensing. Sensors have been developed to measure vibration \cite{kuang2020vibration}, temperature \cite{bhattacharjee2021material}, or pressure inside a fluid-filled cavity \cite{fishel2012sensing, truby2018soft}. To enable safe operation around humans, some collaborative robots (cobots) have been constructed with actuator torque sensors \cite{bischoff2010kuka, petrea2021interaction}. Other research has used the compliant joints of a robot for force estimation \cite{koonjul2011measuring}.

Six-axis force/torque sensors are ubiquitous in both research and industrial applications. These sensors function by measuring elastic deformation in the structure of the sensor \cite{cao2021six}. Force/torque sensors are often placed between the robot arm and end-effector, and they have been used for a wide array of applications such as robotic surgery \cite{puangmali2008state, trejos2010force}%, kim2017surgical}
, humanoid robot locomotion \cite{wu2011optimum}, cloth manipulation \cite{estevez2017robotic}, and robot-assisted dressing and feeding \cite{yu2017haptic, park2020active}. Force/torque sensors are also widely used in industrial applications, including sanding \cite{maric2020collaborative}, deburring \cite{pires2002force}, part alignment \cite{tang2016autonomous}, and robot teaching \cite{kushida2001human}.

\subsection{Tactile Sensing using Vision}

We make the distinction between two types of vision-based approaches to understand tactile signals, those using \textit{internal} versus \textit{external} sensors.

Researchers have developed tactile sensors that use a camera mounted \textit{inside} a robot to perceive a soft exterior. Prior techniques have used dot patterns \cite{ward2018tactip, kuppuswamy2020soft, yamaguchi2016combining}, photometric stereo \cite{yuan2017gelsight}, or fiducial markers \cite{9196925}
to track the motion of the exterior and infer deflection at a high resolution. A variety of other optical techniques have been proposed \cite{shimonomura2019tactile} to sense deformation from internal sensors.

% Many approaches have been developed using \textit{internal} cameras mounted inside the robot to perceive contact. In general, the cameras are mounted 
% Deflections in the soft fingertip surface can be measured using an \textit{internal} camera \cite{yuan2017gelsight, ward2018tactip, kuppuswamy2020soft, yamaguchi2016combining, shimonomura2019tactile} to infer deformation at high resolution. 

Other work uses cameras mounted \textit{external} to the robot to estimate forces. A common technique is to leverage the deformation caused by a rigid gripper making contact with soft objects. This approach has seen success for estimating forces on soft tissue during surgery \cite{kennedy2005vision, nazari2021image}%, chua2021toward}
  ~and manipulation of soft household objects \cite{kim2019efficient}. Other work has used the trajectory of grasped objects to infer the net forces and moments necessary to cause this motion \cite{pham2017hand, li2019estimating}.

The deformation caused by a soft body interacting with a rigid environment can also be used to perceive force for human hands and bodies \cite{grady2022pressurevision, clever2022bodypressure}. Urban \etal~\cite{urban2013computing} use a camera mounted to a human fingernail to observe changes in appearance to estimate force and torque.

This paper builds upon work for estimating the contact pressure applied by a soft gripper to a planar surface \cite{vpec}. They use external static cameras in a \textit{controlled} environment to observe deformations in the part of the gripper contacting the surface. Our work estimates forces and torques measured at the base of the gripper in \textit{uncontrolled} environments and only uses sensors mounted to the robot. This allows collection of a more diverse training dataset and contributes to \network{}'s ability to generalize to novel environments.

% \subsection{Soft Robotic Grippers}

\section{METHODS}
In this section, we describe the grippers used in this study, the data collection hardware and protocol, and the network architecture and training process for \network{}.

\subsection{Selected Grippers}

In order for the forces and torques exerted by a gripper to be visually detected, the gripper must visibly change under the application of force. As a result, we focus on \textit{soft grippers}. These grippers are compliant by design, and this compliance enables the visual estimation of forces. We demonstrate our method with two models of gripper:

\subsubsection{Tendon-Actuated Gripper} The first gripper used is a tendon-actuated gripper included with the Stretch RE1 robot (Figure 1). The gripper features rubber suction-cup-like fingertips which are supported by spring steel flexures. The robot closes the gripper by retracting the inner flexures, causing the fingertips to be pulled together.

During the application of forces and torques, the flexures deform sightly, causing a noticable displacement in the fingertips of the gripper. Figure \ref{fig:axes}b shows the response of the gripper to a load on one of the fingertips. The rubber fingertips are soft, and parts of the gripper may conform to the object being manipulated, especially under high force.

\subsubsection{Pneumatic Gripper} The second gripper used is a commercially-available pneumatic gripper manufactured by SoftGripping \cite{softgripping}. This gripper is made of a flexible silicone rubber, and is constructed with a hollow cavity in each finger. The finger can be filled with compressed air, causing an asymmetric bending and closure of the gripper.
%As this gripper is more compliant than the \textit{Stretch Gripper}, the forces exerted are lower.

During the application of forces and torques, the entire silicone finger deforms relatively uniformly (Figure \ref{fig:axes}c).
%Due to the homogeneous structure of the gripper, we observe that deformations are more isotropic than with tendon-actuated gripper.
Generally, the gripper is less stiff than the tendon-actuated gripper and large deflections of the fingertips are possible under high forces.

% During data collection, the pneumatic gripper was not mounted to a robot, and was instead mounted to a long handle such that the gripper could be operated by a human.

\subsection{Data Capture Hardware}
During collection of the training and test datasets, we mount a 6-DoF ATI Mini45 force/torque sensor \cite{ATI} between the gripper and the wrist of the robot (Figure 1). This allows the accurate sensing of forces and torques applied to the gripper. We do not perform any transforms on the force/torque data from the sensor. Since the sensor drifts over the course of tens of minutes, before each recording, we set the gripper to a level pose and zero the force/torque sensor readings. As a consequence, \network{} estimates the forces and torques applied to the wrist of the gripper, minus the gripper's weight when it is level.

\begin{figure*}
  \centering
  % \vspace{2mm}
  \includegraphics[width=0.95\linewidth]{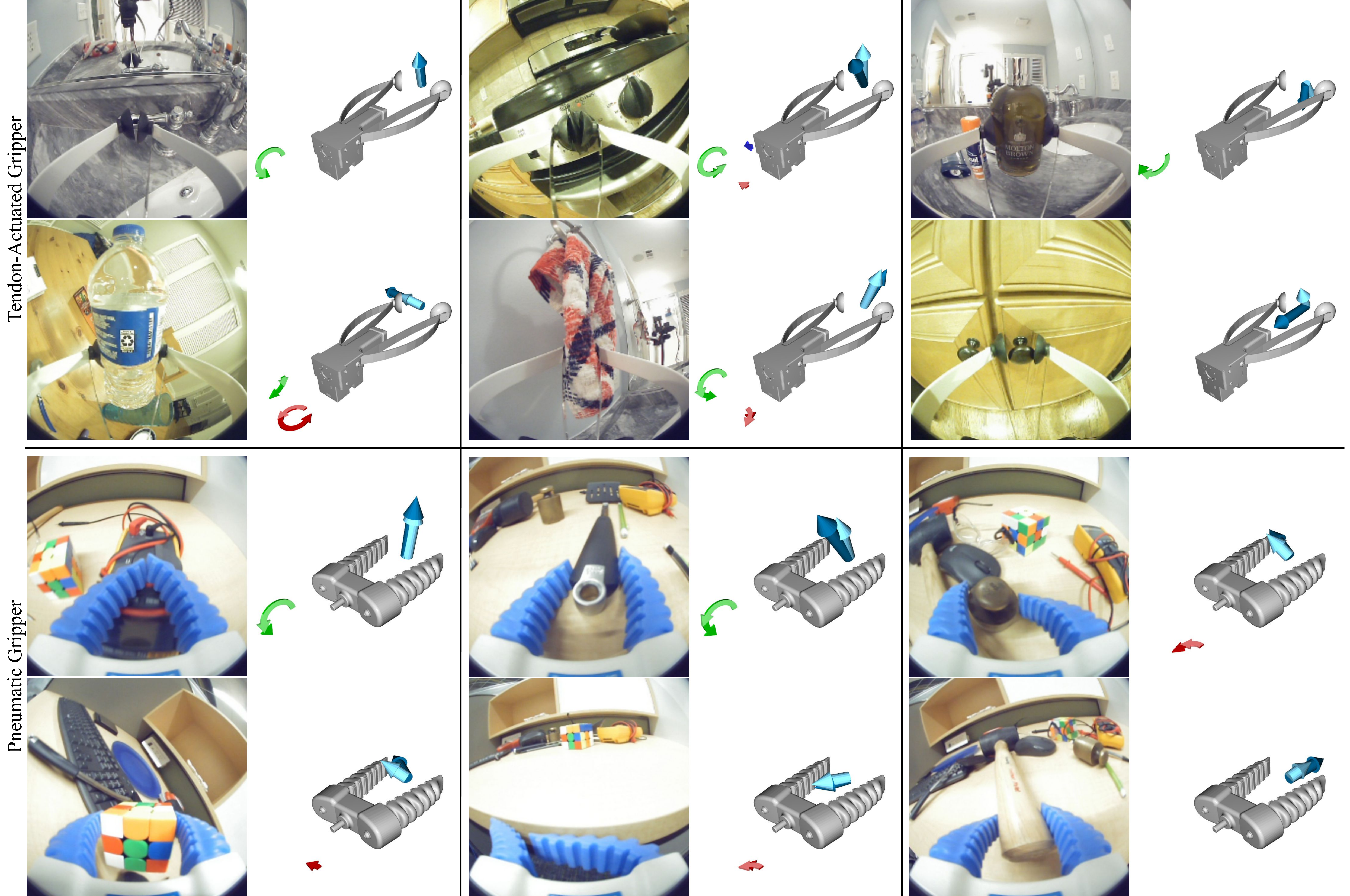}
  % \vspace{-5mm}
  \caption{Frames from the test set for both the tendon-actuated gripper and the pneumatic gripper are shown. Estimates from \network{} are visualized as lightly shaded arrows, and ground truth measurements from the force/torque sensor are darkly shaded arrows. We collected data while the gripper was interacting with objects in a variety of environments, including a bathroom, kitchen, dining room, and office space. The bottom right figure for both gripper types shows a failure case. Both grippers estimate force in the Z-axis less accurately.}
  \label{fig:collage}
\end{figure*}

To capture images of the gripper, we mount a Spinel UC20MPG-F185 fisheye camera above the gripper (Figure \ref{fig:axes}) using 3D-printed hardware. Both the camera and the mount are a part of the \textit{Stretch Teleop Kit} from Hello Robot Inc. The camera captured images at 30 FPS and 640x480 resolution. Due to the wide-angle lens, we crop the images to focus on the area around the gripper (Figure \ref{fig:collage}). We also synchronize the camera and force/torque sensor readings in time.

% To capture images of the gripper, we mounted a Spinel UC20MPG-F185 fisheye camera above the gripper (Figure \ref{fig:axes}) using the open hardware \textit{Stretch Teleop Kit} design released by Hello Robot Inc. The camera captured images at 30 FPS and 640x480 resolution. Due to the wide-angle lens we cropped the images to focus on the area around the gripper (Figure \ref{fig:collage}). We also synchronized the camera and force/torque sensor readings in time.

During data collection with the tendon-actuated gripper, we mount the gripper to a Stretch RE1 mobile manipulator robot by Hello Robot Inc. \cite{designofstretch}. The robot was equipped with the \textit{Dexterous Wrist} add-on, allowing full 6-DoF control of the end-effector. We also record the robot state at each time step, allowing the creation of a baseline based on motor currents. During data collection with the pneumatic gripper, we mount the gripper to a long handle to enable manual operation.

\subsection{Data Capture Protocol} \label{sec:data_protocol}

% \begin{figure}
%   \centering
%   % \vspace{2mm}
%   \includegraphics[width=0.6\linewidth]{images/stretch_in_house.jpg}
%   % \vspace{-5mm}
%   \caption{Data was collected for the tendon-actuated gripper by teleoperating the robot in a variety of settings, including a real home. The robot interacted with objects (e.g. toothbrush, shaving cream) and the environment (e.g. light switch, faucet handle).}
%   \label{fig:robotinhome}
% \end{figure}

We collect approximately 2 hours of data for each gripper. Our approach does not require external sensors, such as cameras mounted to the environment, which simplifies collecting data in a variety of settings. We collect data with each gripper in a lab setting and two distinct office settings. For the tendon-actuated gripper, we also collect data in a home environment. For each gripper, we hold out one complete environment for testing. For the tendon-actuated gripper we hold out the home environment, and for the pneumatic gripper we hold out an office environment. 

%\jeremy{potential hole: reviewers might claim that the soft gripper performed better because one of the office environments is in the training set and another office environment is in the test set}

To capture varied gripper actions, we collect three primitive interaction types: \textit{push}, in which one or both fingertips of the gripper are pushed into a surface; \textit{slide}, in which one or both fingertips of the gripper are dragged along a surface; and \textit{grasp}, in which objects of varying size and mass were grasped at different orientations. In addition, we collect two high-level interaction types: \textit{manipulate object}, which includes tool use, multi-object interactions, and using objects for their intended functions, and \textit{manipulate scene}, which includes interactions with in-context items such as drawers, light switches, and faucets. We vary the wrist roll, wrist pitch, wrist yaw, and grip force throughout the recording process. We record at least 150 seconds of data for each combination of action and environment.

% Two high-level actions are also included in the training and testing set, which are combinations of the three interaction primitives: \textit{object manipulate}, in which objects are grasped and manipulated, and \textit{environment manipulate}, in which a variety of objects in each environment are interacted with, e.g. drawers, light switches, and faucets. The roll, pitch, yaw, and grip are varied throughout the recording process. At least 150 seconds of data is recorded for each combination of action and environment. \patrick{jeremy, can this be simplified?}

%the force/torque sensor was calibrated before each trial by setting the roll, pitch, yaw, and gripper position to a calibration position before zeroing all ground truth forces and torques to account for the effects of gravity.

\subsection{Network Architecture}
We designed \network{} to estimate forces and torques given a single RGB image of the gripper. We train one network for each gripper. The network has a ResNet-18 backbone initialized with weights from pretraining on ImageNet \cite{deng2009imagenet}, and we modify the last linear layer of the network to output 6 scalar values (Figure 1) to directly represent forces and torques.

We train the network with a mean-squared error loss to minimize the error between ground truth force and torque vectors $F, T$ and estimated force and torque vectors $\hat{F}, \hat{T}$. 
\begin{align}
    L = ||F - \hat{F}||^2 + c||T - \hat{T}||^2
\end{align}

To compensate for the intrinsic scale difference between force (Newtons) and torque (Newton-meters), we weight the loss for torque with a constant $c$. We set $c$ equal to the ratio of the force and torque standard deviations in the training data for each gripper. 

% To compensate for the intrinsic scale difference between force (Newtons) and torque (Newton-meters), we weighted the loss for torque with a constant $c$. We set $c=5.6$ for the tendon-actuated gripper and $c=9.3$ for the pneumatic gripper, which is equal to the relative standard deviations of force and torque in the training data for each gripper, respectively. 

% \ck{did you perform this augmentation with the motor efforts?} \jeremy{do you mean for the loss function or the horizontal flipping?} \patrick{no, and I don't think it would be wise to do horizontal flips for motor effort} \ck{i was nervous that you did try to flip images and the motor efforts, since it would be easy to go wrong. if you didn't, I think that would be safer although there wouldn't be as much training data}
During training, we horizontally flip images and the corresponding ground truth F/T sensor readings at random. We also augment the data by varying the brightness, contrast, saturation, and hue of the training images.

We train \network{} for 300k iterations with a learning rate of $1e\text{-}4$ and a batch size of 4. We use the PyTorch framework \cite{pytorch} and the Adam optimizer \cite{adam} to train the network. During evaluation, the network runs at 120 FPS on a desktop PC with an NVIDIA GeForce RTX 3090 GPU, or 25 FPS on the Stretch RE1's Intel i5-8259U CPU.

\begin{table}
\centering
% use resizebox if table gets too big, also uncomment the } after \end{tabular}
%\resizebox{\textwidth}{!}{
\begin{tabular}{c|c|c|c}
    \textbf{Method} & \textbf{Gripper} & \textbf{$RMSE_F$ (N)} & \textbf{$RMSE_T$ (Nm)} \\ \hline
    % Zero-Guesser & Tendon-Actuated & 2.550 & 0.419\\\hline
    Mean-Guesser & Tendon-Act. & 2.405 & 0.389\\\hline
    Effort Baseline & Tendon-Act. & 1.936 & 0.246\\\hline
    \textbf{\network{}} & Tendon-Act. & \textbf{1.688} & \textbf{0.185}\\\hline\hline
    % Zero-Guesser & Pneumatic & 1.468 & 0.158 \\\hline
    Mean-Guesser & Pneumatic & 1.388 & 0.144 \\\hline
    \textbf{\network{}} & Pneumatic & \textbf{0.808} & \textbf{0.051} \\\hline
\end{tabular}
%}
\caption{Force and Torque Error on Test Set
\label{tab:test_set_error}}
\end{table}

\begin{figure*}
  \centering
  % \vspace{2mm}
  \includegraphics[width=1.0\linewidth]{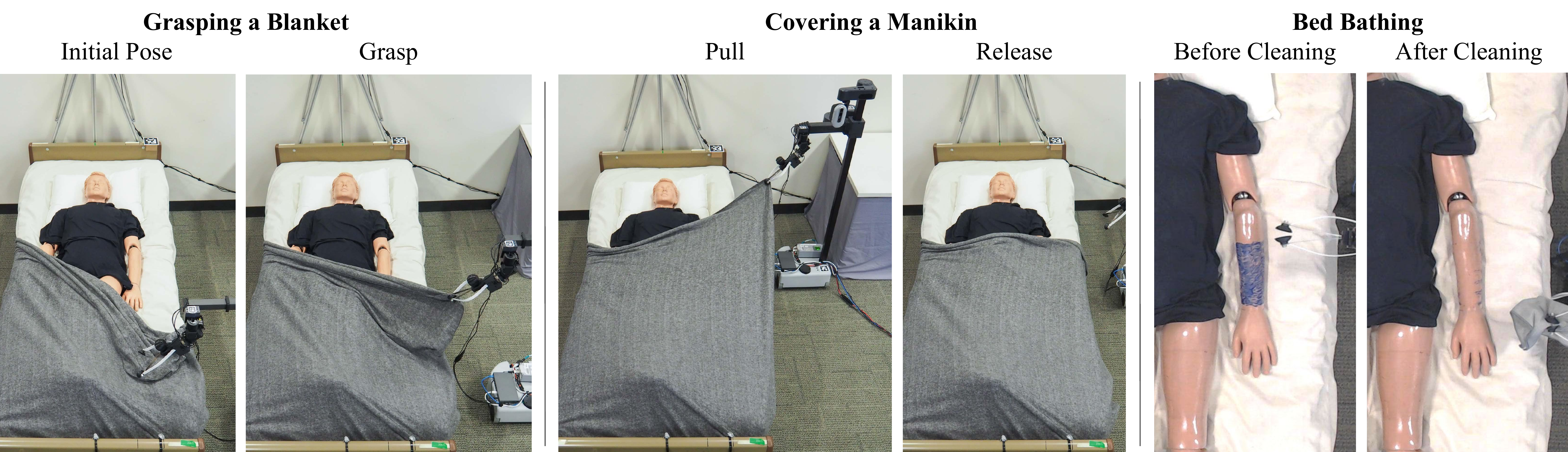}
  % \vspace{-5mm}
  \caption{\textbf{Left:} In the blanket grasping task, the robot travels downward onto a blanket until a vertical force threshold is reached, then the blanket is grasped. \textbf{Center:} In the manikin covering task, the robot carries the corner of a blanket towards the top of the bed until the blanket pulls taught. The blanket is released once a force threshold is exceeded. \textbf{Right:} In the bed bathing task, the robot wipes dry-erase marker off of a manikin using \network{} to regulate force and follow the curvature of the manikin body.}
  \label{fig:cloth}
\end{figure*}
\begin{figure}
  \centering
  % \vspace{2mm}
  \includegraphics[width=1.0\linewidth]{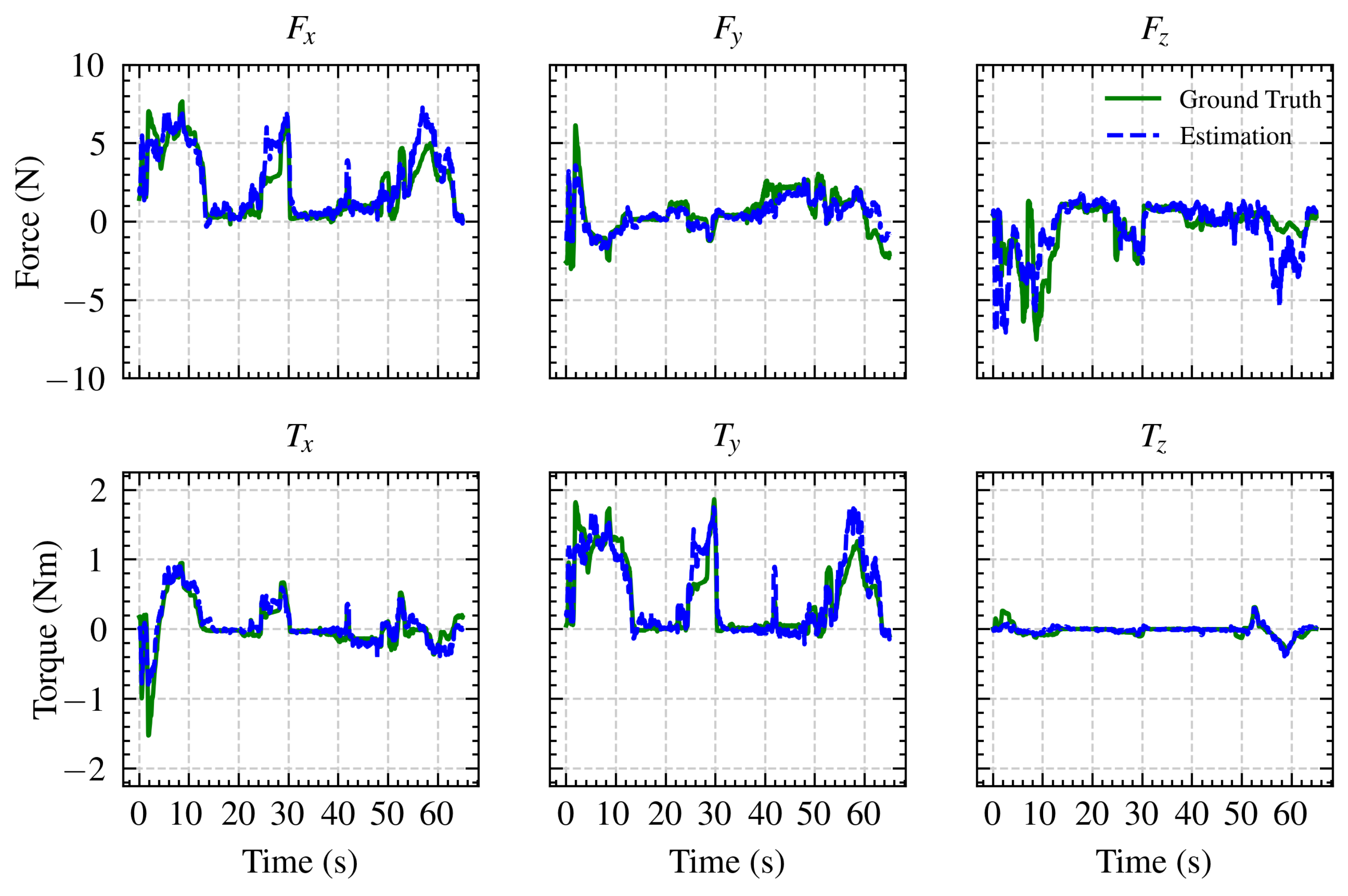}
  % \vspace{-5mm}
  \caption{Typical force and torque estimates for a data sequence recorded with the tendon-actuated gripper. Generally, the estimates from \network{} and the force/torque sensor show good temporal alignment.}
  \label{fig:ft_vs_time_9_6}
\end{figure}

\section{Evaluation} \label{sec:evaluation}

We conducted a variety of experiments to assess the performance of force and torque estimation. We report results with the held-out test data. In addition, we present three evaluations with a mobile manipulator to demonstrate the relevance of \method{} to real-world tasks.

%where force/torque sensors have traditionally been used.

  % In the fabric manipulation task, the robot begins the trial with the blanket edge in the gripper. The robot travels towards the top of the bed until the blanket pulls taught and a force threshold is exceeded. The blanket is then lowered and released. 

\subsection{Evaluation on Held-Out Test Set} We evaluated the performance of \network{} on the test set to assess its ability to infer the forces and torques reported by a conventional wrist-mounted force/torque sensor. In Table \ref{tab:test_set_error} we report the root-mean-squared error for estimated force and torque compared to the ground truth force and torque with force RMSE calculated as:
\begin{align}
    RMSE_F = \sqrt{\frac{1}{N}\sum||\hat{F}-F||^2}
\end{align}
%where $N$ is the number of datapoints in the test set.

%$N=$ 65k for the tendon-actuated gripper and 40k for the soft gripper.\\
%\ck{it would be good to quantify the test sets in terms of the number of frames, the length of time, and any other notable properties}

%compared to . This metric captures the accuracy of the force and torque estimates, $\hat{F}$ and $\hat{T}$, compared to ground truth. Force RMSE can be calculated as:

%We report two types of error metrics computed on the test set in Table \ref{tab:test_set_error}.
%The error is computed for each axis individually, as well as the euclidean error for 3-axis forces and 3-axis torques. \patrick{explain the weighted cosine similarity}

% \begin{itemize}
%     \item \textbf{Mean Squared Force/Torque Error}: To evaluate the accuracy of the estimates, the mean squared error between the ground truth and estimated vectors for force and torque is calculated. Force error is calculated with:

%     \begin{align}
%         MSE_F = \frac{1}{N}\sum||\hat{F}-F||^2
%     \end{align}
    
%     \item \textbf{Weighted Cosine Similarity}: To evaluate the angular alignment between the estimated and ground truth vectors, we calculate the weighted cosine similarity using the angle between the vectors $\theta$. Each sample is weighted by the ground truth vector magnitude.

%     \begin{align}
%         WCS_F = \frac{\sum cos\theta||F||}{\sum||F||}
%     \end{align}
    
% \end{itemize}

We also consider two baselines for comparison against \network{}:
\begin{itemize}
    \item \textbf{Mean-Guesser}: We report the error when always guessing the mean forces and torques from the training set. 
    \item \textbf{Effort Baseline}: Since motor current relates to output torque, motor currents can be used to approximate the load on the gripper. The Stretch RE1 does not directly report motor current, but provides effort, a unitless quantity proportional to current. We train a multilayer perceptron (MLP) with 2 hidden layers to estimate force and torque using the same loss function, ground truth data, and testing set as \network{}. The input to the network is a vector of six efforts from the following actuators: wrist roll, wrist pitch, wrist yaw, gripper, telescoping arm, and arm lift. Since we do not use the robot to collect data with the pneumatic gripper, we only report results on the effort baseline for the tendon-actuated gripper.
    % Since we only used the robot to collect data for the tendon-actuated gripper, we only report results on the effort baseline for it.
\end{itemize}

% \begin{itemize}
%     \item \textbf{Force and Torque Error:} We compute the Euclidean distance between the estimated and ground truth force and torque vectors.
%     \begin{align}
%         E_F = ||F-\hat{F}||
%     \end{align}
    
    % \item \textbf{Angular Cosine Error:} Both force and torque can be represented as vectors. We compute the cosine distance between the ground truth and estimated vectors. For a significant part of the dataset the gripper is not in contact with a surface, so this metric is only computed when the force and torque are greater than their 50th percentile values.
    % \begin{align}
    %     E_{F*\theta}=\frac{F \cdot \hat{F}}{||F||~||\hat{F}||}
    % \end{align}
% \end{itemize}

% \begin{figure*}
%   \centering
%   % \vspace{2mm}
%   \includegraphics[width=1.0\linewidth]{images/ft_axes_2d_histogram.pdf}
%   % \vspace{-5mm}
%   \caption{Scatter plots of each axis.}
%   \label{fig:axes_scatter}
% \end{figure*}

As shown in Table \ref{tab:test_set_error}, \network{} outperformed the Mean-Guesser and the Effort Baseline. Several frames from the test set are shown in Figure \ref{fig:collage}. Figure \ref{fig:ft_vs_time_9_6} illustrates that the estimated forces and torques have temporal alignment with ground truth measurements.

\begin{figure*}
  \centering
  % \vspace{2mm}
  \includegraphics[width=1.0\linewidth]{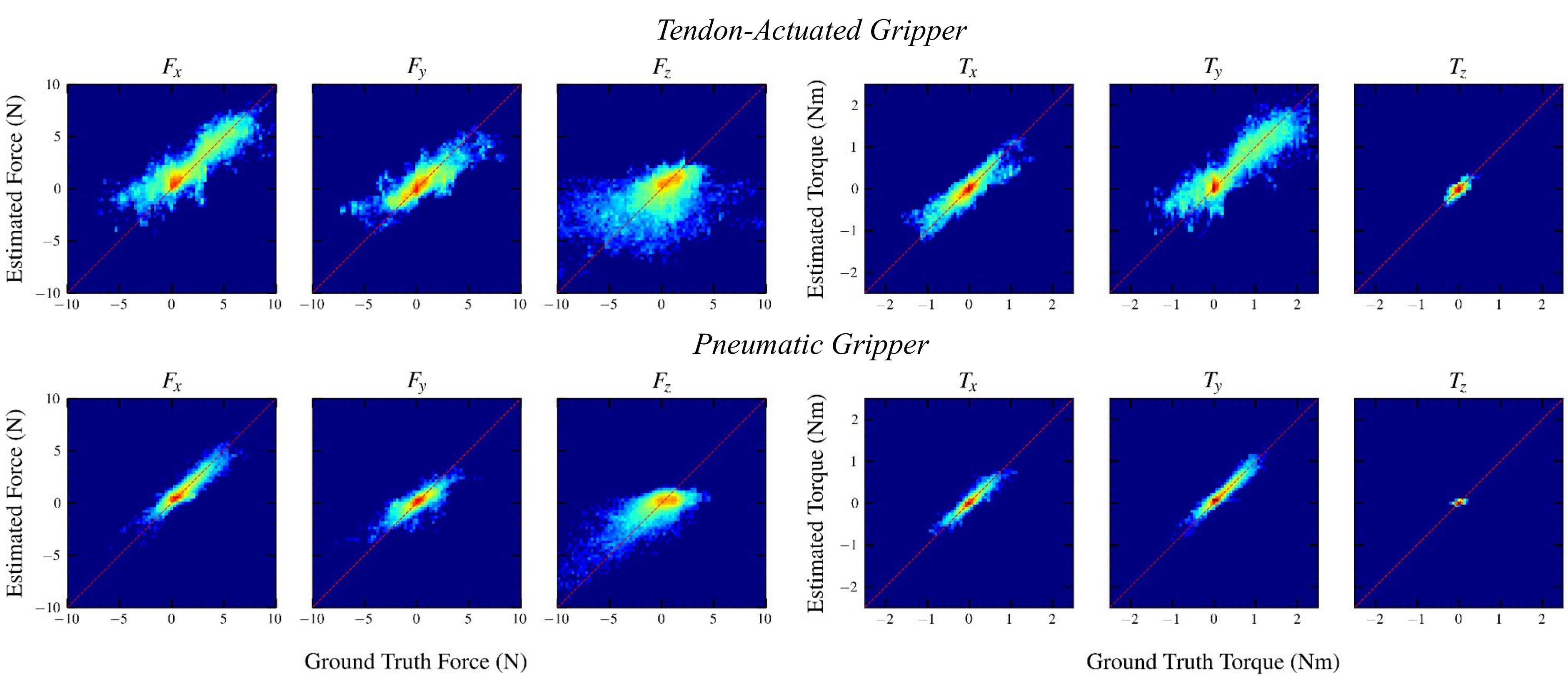}
  \vspace{-5mm}
  \caption{The ground truth and estimated forces and torques across the test set are visualized as a 2D histogram with a logarithmic coloring scale.}
  \label{fig:axes_scatter}
\end{figure*}

\subsection{Grasping a Blanket}
To evaluate the ability of \method{} to support autonomous manipulation, we implemented and tested three primitive fabric manipulation behaviors.
% Prior methods for fabric manipulation have used force-sensing approaches \cite{kudo2000multi, kruse2015collaborative}.

The first task evaluates the robot's ability to stop when a force threshold has been exceeded when picking up a blanket off the surface of a bed. The robot starts with the gripper above the edge of a blanket and then moves downward until the vertical force estimated by \network{} exceeds a 3N threshold. The robot then closes its gripper and picks the blanket up. If the robot successfully grasps and lifts the blanket, we consider the trial to be successful. Table \ref{tab:blanket_results} shows the results.

\subsection{Pulling a Blanket Over a Manikin}

In the second task, the robot pulls a blanket over a manikin. This task evaluates the robot's ability to pull a blanket and stop when a horizontal force threshold is exceeded. The trial begins with the blanket in the gripper, then the mobile base is commanded to travel toward the head of the bed. When the horizontal force estimated by \network{} exceeds a threshold of 2.5N, the robot lowers and releases the blanket. If the blanket was not unintentionally released from the gripper and the robot lowers the blanket onto the manikin, we consider the trial to be successful. Table \ref{tab:blanket_results} shows the results.

\begin{table}
\centering
% use resizebox if table gets too big, also uncomment the } after \end{tabular}
%\resizebox{\textwidth}{!}{
\begin{tabular}{c|c}
    \textbf{Task} & \textbf{Success/Trials}\\ \hline
    Grasp Blanket & 10/10 \\ \hline
    Cover Manikin & 10/10 \\ \hline
\end{tabular}
%}
\caption{Results of Fabric Manipulation Evaluation
\label{tab:blanket_results}}
\end{table}

\subsection{Cleaning a Manikin's Forearm and Leg}

% \begin{table*}
% \centering
% % use resizebox if table gets too big, also uncomment the } after \end{tabular}
% %\resizebox{\textwidth}{!}{
% \begin{tabular}{c|c|c|c|c|c|c|c}
%     \textbf{Method} & \textbf{Gripper} & \textbf{$F_x$ RMSE (N)} & \textbf{$F_y$ RMSE (N)} & \textbf{$F_z$ RMSE (N)} & \textbf{$T_x$ RMSE (Nm)} & \textbf{$T_y$ RMSE (Nm)} & \textbf{$T_z$ RMSE (Nm)} \\\hline
%     Mean Guesser & Tendon-Actuated & 2.331 & 1.562 & 3.079 & 0.278 & 0.613 & 0.049 \\\hline
%     Effort Baseline & Tendon-Actuated & 1.424 & 1.538 & 2.618 & 0.285 & 0.305 & 0.089 \\\hline
%     \textbf{\network{}} & Tendon-Actuated & 1.221 & 0.940 & 2.484 & 0.134 & 0.289 & 0.040\\\hline \hline
%     Mean Guesser & Pneumatic & 1.066 & 0.867 & 1.972 & 0.137 & 0.208 & 0.021 \\\hline
%     \textbf{\network{}} & Pneumatic & 0.440 & 0.530 & 1.219 & 0.055 & 0.067 & 0.019 \\\hline
% \end{tabular}
% %}
% \caption{Axis-Wise Force and Torque Error on Test Set \patrick{I propose omit}
% \label{tab:test_set_axis_error}}
% \end{table*}

Tasks such as cleaning, polishing, and sanding often use force/torque sensing to apply pressure to the surface. We evaluate \method{}'s ability in a similar task of cleaning a medical manikin's limbs. This is related to bathing, which is an important activity of daily living (ADL) with which many people require assistance \cite{mitzner2014identifying}.

%Compared to other activities of daily living (ADLs) and instrumental activities of daily living (iADLs), individuals receiving assistance from caregivers most commonly receive assistance with bathing 

% The ability of a sensor to accomplish such a task demonstrates the ability to quickly determine the direction and magnitude of forces between an end-effector and a surface.

We design a simple controller to clean the limbs of a manikin using force/torque sensing performed by \method{}. The robot starts by holding a cloth above the manikin's limb. It then lowers the gripper until the force magnitude exceeds a 5N threshold, at which point it stops lowering its gripper. The estimated force vector $\hat{F}$ is assumed to be approximately normal to the surface. 

The robot's pose at time $t$ is denoted as $X_t$. The robot moves its gripper tangent to the estimated surface normal toward a desired wiping direction $\vec{d}$. Simultaneously, the robot moves along the estimated force vector $\hat{F}$ to maintain a desired contact force magnitude, $k_f$. Equation \ref{eqn:control_law} represents the control algorithm used by the robot. A new force estimate from \method{} is used at each time step. The robot moves along the limb by 2 cm at each time step, and reverses the desired direction of motion $\vec{d}$ once the gripper has reached a minimum vertical position or has not detected contact with the surface for more than 3 seconds. This results in a back-and-forth motion. The robot does not have any prior knowledge of the cleaning surface geometry.

% The robot's pose at time $t$ is denoted as $X_t$. The robot moves its gripper tangent to the estimated surface normal toward a desired wiping direction $\vec{d}$. Simultaneously, the robot moves along the vector $\hat{F}$ to maintain a desired contact force magnitude, $k_f$. Equation \ref{eqn:control_law} represents the control algorithm used by the robot. A new force estimate from \method{} is used at each time step. The robot reverses the desired direction of motion $\vec{d}$ and moves along the limb by 2 cm once the gripper has reached a minimum vertical position or it has not detected contact with the surface for more than 3 seconds. This results in a back-and-forth motion. The robot does not have any prior knowledge of the cleaning surface geometry.

\begin{align}
  X_{t+1} = X_{t} + (||\hat{F}|| - k_F)\frac{\hat{F}}{||\hat{F}||} + \vec{d}-proj_{\hat{F}}(\vec{d})
  \label{eqn:control_law}
\end{align}

\begin{table}[b]
\centering
% use resizebox if table gets too big, also uncomment the } after \end{tabular}
%\resizebox{\textwidth}{!}{
\begin{tabular}{c|c|c}
    \textbf{Region} & \textbf{Trials} & \textbf{Mean Cleaned Area} \\ \hline
    Arm & 5 & 87.7\% \\\hline
    Leg & 5 & 87.5\% \\\hline
\end{tabular}
%}
\caption{Results of Cleaning a Manikin's Limbs
\label{tab:bathing_results}}
\end{table}

% \begin{figure}
%   \centering
%   % \vspace{2mm}
%   \includegraphics[width=0.5\linewidth,angle=90]{images/manikin.jpg}
%   % \vspace{-5mm}
%   \caption{Overview of Bed-Bathing task with before/after}
%   \label{fig:manikin}
% \end{figure}

% \jeremy{Insert diagram showing normal and tangent deltas}

% $$ \vec{\Delta_{tan}} = (\vec{d_{des}} - \frac{\vec{d_{des}} \cdot \hat{\vec{F}}}{||\hat{\vec{F}}||^2} \hat{\vec{F}})\Delta_{des}$$

% $$ \vec{\Delta_{norm}} = \pm \frac{\hat{\vec{F}}}{||\hat{\vec{F}}||} \Delta_{des} $$

% $$ \begin{bmatrix}
%     x_{t+1} \\
%     y_{t+1} \\
%     z_{t+1}
% \end{bmatrix} = \begin{bmatrix}
%     x_t \\
%     y_t \\
%     z_t
% \end{bmatrix} + \vec{\Delta_{norm}} + \vec{\Delta_{tan}}$$

% \begin{figure}[t]
%   \centering
%   % \vspace{2mm}
%   \includegraphics[width=0.6\linewidth]{images/curved_surface_illustration.pdf}
%   % \vspace{-5mm}
%   \caption{For the manikin cleaning task, the gripper moves along the plane normal to the estimated force to move tangent to the surface.}
%   \label{fig:curved_surface_explanation}
% \end{figure}

Using this control algorithm, the robot cleans off dry-erase marker from the limb of a manikin in bed. We conduct 10 trials: 5 trials on the manikin's left forearm and 5 trials on the manikin's left thigh. No parameters were varied or tuned during the trials aside from the starting position. We define success as the percentage of the marker area erased as measured using a simple color filter applied to images taken by an overhead camera. In our 10 trials, the robot cleaned an average of 87\% of the marker off of the manikin's limbs (see Table \ref{tab:bathing_results}). The algorithm typically succeeded near the center of the arm and leg, but failed on the edges of the leg due to the high curvature and inaccuracies in the Z-axis force estimation.

\subsection{Discussion of Performance}
Figure \ref{fig:axes_scatter} shows 2D histograms calculated by comparing the estimated and ground truth force and torque for each of the six axes. The figure shows the ranges of forces and torques in the testing dataset and visualizes per-axis performance.% Due to the mounting of the force/torque sensor, a shorter moment arm is provided when applying torque around the Z-axis as compared to the X and Y axes. As a result, the magnitude of applied Z-axis torques is lower.

We found that for both grippers, force estimation in the Z-axis performs more poorly than the X and Y axes. We suspect that this is because deformations in depth are less observable with a single RGB image, and more likely to be obscured by held objects and gripper forces. In contrast, forces in the X and Y axes are aligned with the image plane. 

The pneumatic gripper estimates have lower error than the tendon-actuated gripper for all axes, which may be due to larger deformations of the pneumatic fingers under load. 

\section{CONCLUSION}

%We present \method{}, a method which uses an eye-in-hand camera to capture the deformation in soft grippers and estimate the applied forces and torques associated with these deformations. We collect data in a variety of environments for two grippers: a tendon-actuated gripper and a pneumatic gripper. We demonstrate that our method can be used to reliably estimate force and torque on a held-out test set. Additionally, the estimates can be applied to unseen scenarios and used to control complex manipulation tasks.

We presented Visual Force/Torque Sensing (\method{}), a method that visually estimates the 6-axis force/torque measurement that would be reported by a conventional wrist-mounted force/torque sensor. Our method uses an external camera with a fisheye lens to observe a soft gripper as it deforms due to loads. \method{} uses a deep learning model that outputs a 6-axis F/T estimate when given a single RGB image as input. Our method outperforms motor-current-based F/T estimation when tested with data from teleoperated manipulation of novel objects in a novel environment. Our method also enabled successful autonomous mobile manipulation of fabrics in three tasks. Overall, our results suggest that an external camera observing a soft gripper can perform useful visual force/torque sensing for a variety of manipulation tasks.

% for soft grippers 

% that would be reported by a

% conventional F/T sensor. In contrast to approaches that sense loads using internal cameras placed behind soft exterior surfaces, our approach uses an external camera with a fisheye lens that observes a soft gripper. \method{} includes a deep learning model that takes a single RGB image as input and outputs a 6-axis F/T estimate. We trained the model with sensor data collected while teleoperating a robot (Stretch RE1 from Hello Robot Inc.) to perform manipulation tasks. \method{} outperformed F/T estimates based on motor currents, generalized to a novel home environment, and supported three autonomous tasks relevant to healthcare: grasping a blanket, pulling a blanket over a manikin, and cleaning a manikin's limbs. \method{} also performed well with a manually operated pneumatic gripper. Overall, our results show that useful F/T estimates can be obtained from visual observations of soft grippers, and \method{} provides a feasible alternative to conventional wrist-mounted F/T sensors and motor-current F/T estimation. 

\bibliographystyle{IEEEtran}
\bibliography{cited}

\begin{thebibliography}{10}
\providecommand{\url}[1]{#1}
\csname url@rmstyle\endcsname
\providecommand{\newblock}{\relax}
\providecommand{\bibinfo}[2]{#2}
\providecommand\BIBentrySTDinterwordspacing{\spaceskip=0pt\relax}
\providecommand\BIBentryALTinterwordstretchfactor{4}
\providecommand\BIBentryALTinterwordspacing{\spaceskip=\fontdimen2\font plus
\BIBentryALTinterwordstretchfactor\fontdimen3\font minus
  \fontdimen4\font\relax}
\providecommand\BIBforeignlanguage[2]{{%
\expandafter\ifx\csname l@#1\endcsname\relax
\typeout{** WARNING: IEEEtran.bst: No hyphenation pattern has been}%
\typeout{** loaded for the language `#1'. Using the pattern for}%
\typeout{** the default language instead.}%
\else
\language=\csname l@#1\endcsname
\fi
#2}}

\bibitem{kuang2020vibration}
W.~Kuang, M.~Yip, and J.~Zhang, ``Vibration-based multi-axis force sensing:
  Design, characterization, and modeling,'' \emph{IEEE Robotics and Automation
  Letters}, vol.~5, no.~2, pp. 3082--3089, 2020.

\bibitem{bhattacharjee2021material}
T.~Bhattacharjee, H.~M. Clever, J.~Wade, and C.~C. Kemp, ``Material recognition
  via heat transfer given ambiguous initial conditions,'' \emph{IEEE
  Transactions on Haptics}, vol.~14, no.~4, pp. 885--896, 2021.

\bibitem{fishel2012sensing}
J.~A. Fishel and G.~E. Loeb, ``Sensing tactile microvibrations with the
  biotac—comparison with human sensitivity,'' in \emph{2012 4th IEEE RAS \&
  EMBS International Conference on Biomedical Robotics and Biomechatronics
  (BioRob)}.\hskip 1em plus 0.5em minus 0.4em\relax IEEE, 2012, pp. 1122--1127.

\bibitem{truby2018soft}
R.~L. Truby, M.~Wehner, A.~K. Grosskopf, D.~M. Vogt, S.~G. Uzel, R.~J. Wood,
  and J.~A. Lewis, ``Soft somatosensitive actuators via embedded 3d printing,''
  \emph{Advanced Materials}, vol.~30, no.~15, p. 1706383, 2018.

\bibitem{bischoff2010kuka}
R.~Bischoff, J.~Kurth, G.~Schreiber, R.~Koeppe, A.~Albu-Sch{\"a}ffer, A.~Beyer,
  O.~Eiberger, S.~Haddadin, A.~Stemmer, G.~Grunwald, \emph{et~al.}, ``The
  kuka-dlr lightweight robot arm - a new reference platform for robotics
  research and manufacturing,'' in \emph{ISR 2010 (41st International Symposium
  on Robotics) and ROBOTIK 2010 (6th German Conference on Robotics)}.\hskip 1em
  plus 0.5em minus 0.4em\relax VDE, 2010, pp. 1--8.

\bibitem{petrea2021interaction}
R.~A.~B. Petrea, M.~Bertoni, and R.~Oboe, ``On the interaction force sensing
  accuracy of franka emika panda robot,'' in \emph{IECON 2021--47th Annual
  Conference of the IEEE Industrial Electronics Society}.\hskip 1em plus 0.5em
  minus 0.4em\relax IEEE, 2021, pp. 1--6.

\bibitem{koonjul2011measuring}
G.~S. Koonjul, G.~J. Zeglin, and N.~S. Pollard, ``Measuring contact points from
  displacements with a compliant, articulated robot hand,'' in \emph{2011 IEEE
  International Conference on Robotics and Automation}.\hskip 1em plus 0.5em
  minus 0.4em\relax IEEE, 2011, pp. 489--495.

\bibitem{cao2021six}
M.~Y. Cao, S.~Laws, and F.~R. y~Baena, ``Six-axis force/torque sensors for
  robotics applications: A review,'' \emph{IEEE Sensors Journal}, 2021.

\bibitem{puangmali2008state}
P.~Puangmali, K.~Althoefer, L.~D. Seneviratne, D.~Murphy, and P.~Dasgupta,
  ``State-of-the-art in force and tactile sensing for minimally invasive
  surgery,'' \emph{IEEE Sensors Journal}, vol.~8, no.~4, pp. 371--381, 2008.

\bibitem{trejos2010force}
A.~Trejos, R.~Patel, and M.~Naish, ``Force sensing and its application in
  minimally invasive surgery and therapy: A survey,'' \emph{Proceedings of the
  Institution of Mechanical Engineers, Part C: Journal of Mechanical
  Engineering Science}, vol. 224, no.~7, pp. 1435--1454, 2010.

\bibitem{wu2011optimum}
B.~Wu, J.~Luo, F.~Shen, Y.~Ren, and Z.~Wu, ``Optimum design method of
  multi-axis force sensor integrated in humanoid robot foot system,''
  \emph{Measurement}, vol.~44, no.~9, pp. 1651--1660, 2011.

\bibitem{estevez2017robotic}
D.~Estevez, J.~G. Victores, R.~Fernandez-Fernandez, and C.~Balaguer, ``Robotic
  ironing with 3d perception and force/torque feedback in household
  environments,'' in \emph{2017 IEEE/RSJ International Conference on
  Intelligent Robots and Systems (IROS)}.\hskip 1em plus 0.5em minus
  0.4em\relax IEEE, 2017, pp. 6484--6489.

\bibitem{yu2017haptic}
W.~Yu, A.~Kapusta, J.~Tan, C.~C. Kemp, G.~Turk, and C.~K. Liu, ``Haptic
  simulation for robot-assisted dressing,'' in \emph{2017 IEEE International
  Conference on Robotics and Automation (ICRA)}.\hskip 1em plus 0.5em minus
  0.4em\relax IEEE, 2017, pp. 6044--6051.

\bibitem{park2020active}
D.~Park, Y.~Hoshi, H.~P. Mahajan, H.~K. Kim, Z.~Erickson, W.~A. Rogers, and
  C.~C. Kemp, ``Active robot-assisted feeding with a general-purpose mobile
  manipulator: Design, evaluation, and lessons learned,'' \emph{Robotics and
  Autonomous Systems}, vol. 124, p. 103344, 2020.

\bibitem{maric2020collaborative}
B.~Maric, A.~Mutka, and M.~Orsag, ``Collaborative human-robot framework for
  delicate sanding of complex shape surfaces,'' \emph{IEEE Robotics and
  Automation Letters}, vol.~5, no.~2, pp. 2848--2855, 2020.

\bibitem{pires2002force}
J.~N. Pires, J.~Ramming, S.~Rauch, and R.~Ara{\'u}jo, ``Force/torque sensing
  applied to industrial robotic deburring,'' \emph{Sensor Review}, 2002.

\bibitem{tang2016autonomous}
T.~Tang, H.-C. Lin, Y.~Zhao, W.~Chen, and M.~Tomizuka, ``Autonomous alignment
  of peg and hole by force/torque measurement for robotic assembly,'' in
  \emph{2016 IEEE International Conference on Automation Science and
  Engineering (CASE)}.\hskip 1em plus 0.5em minus 0.4em\relax IEEE, 2016, pp.
  162--167.

\bibitem{kushida2001human}
D.~Kushida, M.~Nakamura, S.~Goto, and N.~Kyura, ``Human direct teaching of
  industrial articulated robot arms based on force-free control,''
  \emph{Artificial Life and Robotics}, vol.~5, no.~1, pp. 26--32, 2001.

\bibitem{ward2018tactip}
B.~Ward-Cherrier, N.~Pestell, L.~Cramphorn, B.~Winstone, M.~E. Giannaccini,
  J.~Rossiter, and N.~F. Lepora, ``The tactip family: Soft optical tactile
  sensors with 3d-printed biomimetic morphologies,'' \emph{Soft Robotics},
  vol.~5, no.~2, pp. 216--227, 2018.

\bibitem{kuppuswamy2020soft}
N.~Kuppuswamy, A.~Alspach, A.~Uttamchandani, S.~Creasey, T.~Ikeda, and
  R.~Tedrake, ``Soft-bubble grippers for robust and perceptive manipulation,''
  in \emph{2020 IEEE/RSJ International Conference on Intelligent Robots and
  Systems (IROS)}.\hskip 1em plus 0.5em minus 0.4em\relax IEEE, 2020, pp.
  9917--9924.

\bibitem{yamaguchi2016combining}
A.~Yamaguchi and C.~G. Atkeson, ``Combining finger vision and optical tactile
  sensing: Reducing and handling errors while cutting vegetables,'' in
  \emph{2016 IEEE-RAS 16th International Conference on Humanoid Robots
  (Humanoids)}.\hskip 1em plus 0.5em minus 0.4em\relax IEEE, 2016, pp.
  1045--1051.

\bibitem{yuan2017gelsight}
W.~Yuan, S.~Dong, and E.~H. Adelson, ``Gelsight: High-resolution robot tactile
  sensors for estimating geometry and force,'' \emph{Sensors}, vol.~17, no.~12,
  p. 2762, 2017.

\bibitem{9196925}
R.~Ouyang and R.~Howe, ``Low-cost fiducial-based 6-axis force-torque sensor,''
  in \emph{2020 IEEE International Conference on Robotics and Automation
  (ICRA)}, 2020, pp. 1653--1659.

\bibitem{shimonomura2019tactile}
K.~Shimonomura, ``Tactile image sensors employing camera: A review,''
  \emph{Sensors}, vol.~19, no.~18, p. 3933, 2019.

\bibitem{kennedy2005vision}
C.~W. Kennedy and J.~P. Desai, ``A vision-based approach for estimating contact
  forces: Applications to robot-assisted surgery,'' \emph{Applied Bionics and
  Biomechanics}, vol.~2, no.~1, pp. 53--60, 2005.

\bibitem{nazari2021image}
A.~A. Nazari, F.~Janabi-Sharifi, and K.~Zareinia, ``Image-based force
  estimation in medical applications: A review,'' \emph{IEEE Sensors Journal},
  vol.~21, no.~7, pp. 8805--8830, 2021.

\bibitem{kim2019efficient}
D.~Kim, H.~Cho, H.~Shin, S.-C. Lim, and W.~Hwang, ``An efficient
  three-dimensional convolutional neural network for inferring physical
  interaction force from video,'' \emph{Sensors}, vol.~19, no.~16, p. 3579,
  2019.

\bibitem{pham2017hand}
T.-H. Pham, N.~Kyriazis, A.~A. Argyros, and A.~Kheddar, ``Hand-object contact
  force estimation from markerless visual tracking,'' \emph{IEEE Transactions
  on Pattern Analysis and Machine Intelligence}, vol.~40, no.~12, pp.
  2883--2896, 2017.

\bibitem{li2019estimating}
Z.~Li, J.~Sedlar, J.~Carpentier, I.~Laptev, N.~Mansard, and J.~Sivic,
  ``Estimating 3d motion and forces of person-object interactions from
  monocular video,'' in \emph{Proceedings of the IEEE/CVF Conference on
  Computer Vision and Pattern Recognition}, 2019, pp. 8640--8649.

\bibitem{grady2022pressurevision}
P.~Grady, C.~Tang, S.~Brahmbhatt, C.~D. Twigg, C.~Wan, J.~Hays, and C.~C. Kemp,
  ``{PressureVision:} estimating hand pressure from a single {RGB} image,''
  \emph{European Conference on Computer Vision (ECCV)}, 2022.

\bibitem{clever2022bodypressure}
H.~M. Clever, P.~L. Grady, G.~Turk, and C.~C. Kemp, ``Bodypressure-inferring
  body pose and contact pressure from a depth image,'' \emph{IEEE Transactions
  on Pattern Analysis and Machine Intelligence}, vol.~45, no.~1, pp. 137--153,
  2022.

\bibitem{urban2013computing}
S.~Urban, J.~Bayer, C.~Osendorfer, G.~Westling, B.~B. Edin, and P.~Van
  Der~Smagt, ``Computing grip force and torque from finger nail images using
  gaussian processes,'' in \emph{2013 IEEE/RSJ International Conference on
  Intelligent Robots and Systems}.\hskip 1em plus 0.5em minus 0.4em\relax IEEE,
  2013, pp. 4034--4039.

\bibitem{vpec}
P.~Grady, J.~A. Collins, S.~Brahmbhatt, C.~D. Twigg, C.~Tang, J.~Hays, and
  C.~C. Kemp, ``Visual pressure estimation and control for soft robotic
  grippers,'' \emph{2022 IEEE/RSJ International Conference on Intelligent
  Robots and Systems (IROS)}, 2022.

\bibitem{softgripping}
\BIBentryALTinterwordspacing
{SoftGripping by Wegard GmbH}. (2022) {SoftGripping}, the modular design system
  for flexible gripping. [Online]. Available: \url{https://soft-gripping.com/}
\BIBentrySTDinterwordspacing

\bibitem{ATI}
\BIBentryALTinterwordspacing
{ATI Industrial Automation}. (2022) {F/T Sensor:} mini45. [Online]. Available:
  \url{https://www.ati-ia.com/products/ft/ft_models.aspx?id=mini45}
\BIBentrySTDinterwordspacing

\bibitem{designofstretch}
C.~C. Kemp, A.~Edsinger, H.~M. Clever, and B.~Matulevich, ``The design of
  stretch: A compact, lightweight mobile manipulator for indoor human
  environments,'' in \emph{2022 International Conference on Robotics and
  Automation (ICRA)}, 2022, pp. 3150--3157.

\bibitem{deng2009imagenet}
J.~Deng, W.~Dong, R.~Socher, L.-J. Li, K.~Li, and L.~Fei-Fei, ``Imagenet: A
  large-scale hierarchical image database,'' in \emph{2009 IEEE Conference on
  Computer Vision and Pattern Recognition, (CVPR)}.\hskip 1em plus 0.5em minus
  0.4em\relax IEEE, 2009, pp. 248--255.

\bibitem{pytorch}
A.~Paszke, S.~Gross, F.~Massa, A.~Lerer, J.~Bradbury, G.~Chanan, T.~Killeen,
  Z.~Lin, N.~Gimelshein, L.~Antiga, A.~Desmaison, A.~Kopf, E.~Yang, Z.~DeVito,
  M.~Raison, A.~Tejani, S.~Chilamkurthy, B.~Steiner, L.~Fang, J.~Bai, and
  S.~Chintala, ``Pytorch: An imperative style, high-performance deep learning
  library,'' in \emph{Advances in Neural Information Processing Systems 32},
  H.~Wallach, H.~Larochelle, A.~Beygelzimer, F.~d\textquotesingle
  Alch\'{e}-Buc, E.~Fox, and R.~Garnett, Eds.\hskip 1em plus 0.5em minus
  0.4em\relax Curran Associates, Inc., 2019, pp. 8024--8035.

\bibitem{adam}
D.~P. Kingma and J.~Ba, ``Adam: {A} method for stochastic optimization,'' in
  \emph{3rd International Conference on Learning Representations, (ICLR) 2015},
  Y.~Bengio and Y.~LeCun, Eds., 2015.

\bibitem{mitzner2014identifying}
T.~L. Mitzner, T.~L. Chen, C.~C. Kemp, and W.~A. Rogers, ``Identifying the
  potential for robotics to assist older adults in different living
  environments,'' \emph{International Journal of Social Robotics}, vol.~6,
  no.~2, pp. 213--227, 2014.

\end{thebibliography}

\end{document}